\documentclass[conference]{IEEEtran}
\IEEEoverridecommandlockouts
\usepackage{cite}
\usepackage{amsmath,amssymb}
\usepackage{algorithmic}
\usepackage{graphicx}
\usepackage{textcomp}
\usepackage{hyperref}
\usepackage{tikz}
\usepackage[utf8]{inputenc}
\usepackage{listings}
\usepackage[T1]{fontenc}
\usepackage{booktabs}
\usepackage{tabularx}        
\usepackage{booktabs}        
\usepackage{array}           
\usepackage{graphicx}   
\usepackage{adjustbox}  
\usepackage{booktabs}

\usepackage{newunicodechar}      
\usepackage{lmodern}             

\newunicodechar{├}{|} 
\newunicodechar{─}{-}
\newunicodechar{│}{|}
\newunicodechar{└}{`-}
\newunicodechar{⋮}{.} 
\usetikzlibrary{shapes, arrows, positioning}
\tikzstyle{block} = [rectangle, draw,  
    text width=2.5cm, text centered, rounded corners, minimum height=1cm]
\tikzstyle{arrow} = [thick,->,>=stealth]
\usepackage{xcolor}
\def\BibTeX{{\rm B\kern-.05em{\sc i\kern-.025em b}\kern-.08em
    T\kern-.1667em\lower.7ex\hbox{E}\kern-.125emX}}

\begin{document}

\title{EgMM-Corpus: A Multimodal Vision-Language Dataset for Egyptian Culture}
\author{
\IEEEauthorblockN{
Mohamed Gamil\IEEEauthorrefmark{1},
Abdelrahman Elsayed\IEEEauthorrefmark{1},
Abdelrahman Lila\IEEEauthorrefmark{1},
Ahmed Gad\IEEEauthorrefmark{1},\\
Hesham Abdelgawad\IEEEauthorrefmark{1},
Mohamed Aref\IEEEauthorrefmark{1},
Ahmed Fares\IEEEauthorrefmark{1}\IEEEauthorrefmark{2}
}
\IEEEauthorblockA{
\IEEEauthorrefmark{1}Department of Computer Science and Engineering,\\
Egypt-Japan University of Science and Technology (E-JUST), Alexandria 21934, Egypt
}
\IEEEauthorblockA{
\IEEEauthorrefmark{2}Department of Electrical Engineering,\\
Faculty of Engineering at Shoubra, Benha University, Cairo 11629, Egypt
}
\IEEEauthorblockA{
Emails: \texttt{\{mohammed.gamil, abdelrahman.elsayed,  abdelrahman.lila, ahmed.gad,}\\
\texttt{hesham.abdelgawad, mohamed.aref, ahmed.fares\}@ejust.edu.eg}
}
}

\maketitle
\begin{abstract}
Despite recent advances in AI, multimodal culturally diverse datasets are still limited, particularly for regions in the Middle East and Africa. In this paper, we introduce \textbf{EgMM-Corpus}, a multimodal dataset dedicated to Egyptian culture. By designing and running a new data collection pipeline, we collected over 3,000 images, covering 313 concepts across landmarks, food, and folklore. Each entry in the dataset is manually validated for cultural authenticity and multimodal coherence. \textit{EgMM-Corpus} aims to provide a reliable resource for evaluating and training vision-language models in an Egyptian cultural context. We further evaluate the zero-shot performance of  Contrastive Language-Image Pre-training CLIP on \textit{EgMM-Corpus}, on which it achieves 21.2\% Top-1 accuracy and 36.4\% Top-5 accuracy in classification. These results underscore the existing cultural bias in large-scale vision–language models and demonstrate the importance of \textit{EgMM-Corpus} as a benchmark for developing culturally aware models.
\end{abstract}

\section{Introduction}

\subsection{Background}
Data is the driving force behind modern Artificial Intelligence (AI), particularly in vision-language modelling. While large-scale culturally diverse datasets such as \textbf{Dollar Street}~\cite{dollarstreet2022}, and \textbf{MaRVL}~\cite{marvl2021} have enabled progress in mitigating cultural and geographic biases, they remain limited in regional coverage. More recent benchmarks, such as \textbf{CultureVerse}~\cite{cultureverse2024}, reveal that VLMs still underperform in African and Asian contexts. This scarcity of high-quality, multimodal data specific to these regions results in models that fail to accurately interpret African cultures-specifically Egyptian visual and cultural contexts.

\subsection{Problem Statement}
Despite advances in multimodal learning, culturally grounded datasets remain scarce. Current benchmarks lack diversity and consistency in representing regional and cultural elements. For instance, a model may identify an Egyptian dish or attire but fail to associate it with its cultural or social meaning. This limitation hinders applications in cultural education, heritage preservation, and tourism.

\subsection{Significance}
Developing culturally representative datasets is vital for fairness, inclusivity, and accurate contextual understanding in AI. For Egypt, a country with millennia of history and diverse social practices-such datasets can strengthen localized AI applications, foster digital preservation of culture, and promote global research on cultural understanding in multimodal learning.

\subsection{Related Work}
Recent research has emphasized the importance of cultural context in multimodal understanding. Several efforts have sought to mitigate cultural and geographic biases in Vision–Language Models (VLMs) by developing diverse datasets. For example, \textbf{Dollar Street}~\cite{dollarstreet2022} captures socioeconomic diversity across multiple countries through indoor images. More recently, \textbf{CultureVerse}~\cite{cultureverse2024} provides a multimodal benchmark for multicultural understanding, revealing persistent weaknesses in VLM performance on African and Asian contexts.

Other notable works include \textbf{CulturalVQA}~\cite{culturalvqa2024}, which introduces a benchmark for cultural question answering across multiple regions, \textbf{CVQA}~\cite{cvqa2025} that extends this to multilingual and multicultural settings, \textbf{CultureBank}~\cite{culturebank2024}, which compiles a global cultural knowledge base for vision-language modeling, and \textbf{HuMNet}~\cite{humannet2023}, focusing on human-material interactions in cultural heritage. While these datasets advance cultural grounding, they remain limited in regional representation, particularly for the Middle East and North Africa (MENA). No existing benchmark provides broad, multimodal, and culturally contextualized data specifically for Egypt, leaving a critical gap that this work addresses.

\subsection{Existing Egyptian Datasets}
Several Egypt-focused datasets exist on open platforms such as Kaggle, including:
\begin{itemize}
    \item \textbf{Egypt Landmarks Dataset}~\cite{kaggle_eg_landmarks} - 5,391 images covering 219 landmarks.
    \item \textbf{Ancient Egyptian Landmarks}~\cite{kaggle_ancient_landmarks} - visual dataset for monument recognition tasks.
    \item \textbf{Egypt Monuments Dataset}~\cite{kaggle_monuments} - images of various historical monuments (e.g., Bent Pyramid, Ramesses~II).

\end{itemize}
While valuable, these resources are narrow in scope, focusing on individual modalities or specific domains. None provides cross-modal alignment or comprehensive cultural diversity. 

\subsection{Research Gap}
Existing datasets are either small, geographically narrow, or lack cross-modal alignment. Annotations often rely on translations that fail to capture local semantics. Crucially, no dataset currently provides comprehensive coverage of Egypt’s visual, linguistic, and social diversity, limiting the development of culturally aware AI systems.

\subsection{Objectives}
This work presents \textit{EgMM-Corpus: A Multimodal Vision-Language Dataset for Egyptian Culture}, a multimodal dataset representing Egyptian culture through images, text, and metadata. It spans multiple cultural categories, as festivals, landmarks, and social practices serve as a foundation for training and evaluating culturally grounded vision-language models.

\subsection{Proposed Dataset}
The dataset integrates verified cultural archives, open repositories, and original field contributions. Each entry undergoes manual annotation and validation to ensure cultural authenticity, linguistic precision, and multimodal coherence. Compared to prior work, \textit{EgMM-Corpus} offers broader coverage, better balance across modalities, and stronger semantic alignment between visual and textual components.

\subsection{Contributions}
The key contributions of this paper are summarized as follows:
\begin{itemize}
    \item Introduction of \textbf{EgMM-Corpus}, a multimodal dataset dedicated to Egyptian culture.
    \item A rigorous data collection and annotation pipeline ensuring cultural integrity and linguistic quality.
    \item Public release of the dataset to advance research in culturally grounded multimodal AI.
    \item Establishment of \textbf{baseline evaluations} using CLIP ~\cite{clip2021} for zero-shot classification and cross-modal retrieval, providing initial benchmarks for future research on Egyptian cultural understanding.
\end{itemize}

\par This paper addresses the scarcity of culturally comprehensive multimodal datasets by introducing \textit{EgMM-Corpus}, a corpus centered on Egyptian culture. By enhancing fairness, inclusivity, and contextual understanding, this dataset paves the way for vision-language models that respect and represent diverse cultural identities.

\section{Methodology}

\subsection{Data Collection Pipeline}
Our data collection pipeline automates the retrieval and organization of multimodal cultural knowledge from both textual and visual sources. The pipeline now consists of modular components designed to flexibly handle multiple cultural domains, including \textbf{monuments}, \textbf{traditional food}, and \textbf{folklore}. Depending on the input type, the pipeline operates either on a list of manually defined concepts (e.g., monuments) or dynamically retrieves cultural entities using only the \textit{country name} (e.g., for cuisine and folklore).

\subsubsection{Concept Retrieval}
We begin with a curated list of Egyptian monuments and heritage sites, manually selected from publicly available cultural directories and Wikipedia categories. Each concept represents a distinct entity such as a landmark, or historical artifact (e.g., \textit{Abu Simbel Temple}, \textit{The Great Pyramid of Giza}).

For culinary and folklore extensions, the process is fully automated:
\begin{itemize}
    \item \textbf{TasteAtlas Integration:} Using the Playwright automation framework, the pipeline scrapes the TasteAtlas website to retrieve lists of traditional dishes associated with a specific country. Each dish name and corresponding link is extracted dynamically by interacting with the “View More” buttons until all entries are loaded~\cite{tasteatlas}.
    \item \textbf{UNESCO Intangible Heritage Integration:} A similar automation step retrieves folklore, traditional crafts, and performance arts from UNESCO’s Intangible Cultural Heritage database. Each heritage entry is extracted along with its description, country, and classification~\cite{unesco_ich}.
\end{itemize}

\subsubsection{Textual Background Extraction}
For every retrieved concept, textual grounding is collected from two high-quality knowledge sources: \textbf{Wikipedia} and the \textbf{Encyclopaedia Britannica}. These sources were selected for their reliability, coverage, and consistent structure. Using the \texttt{BeautifulSoup}, \texttt{requests}, and \texttt{playwright} libraries, the system automatically retrieves:
\begin{itemize}
    \item Article titles, introductory sections, and descriptive content,
\end{itemize}
The processed text is stored in a Markdown file (\texttt{background.md}) associated with each concept, serving as a unified textual grounding source for subsequent reasoning and question generation tasks.

\subsubsection{Visual Content Collection}
To provide visual grounding, the pipeline queries \textbf{DuckDuckGo Images} for each concept title or dish name. Images are filtered for quality and license availability, then downloaded and stored in a structured directory format. This component ensures a balanced multimodal dataset suitable for visual reasoning and cultural representation tasks.

\subsection{Data Organization and Storage}
Each concept-whether a monument, dish, or folklore element-is represented by a structured directory containing both textual and visual data:

\begin{verbatim}
/concept_id/
    |── 0.jpg
    │── 1.jpg
    │── ...
    ├── background.md
\end{verbatim}

This structure facilitates reproducibility and ensures compatibility with downstream tasks such as question answering, image captioning, and knowledge-grounded multimodal reasoning.





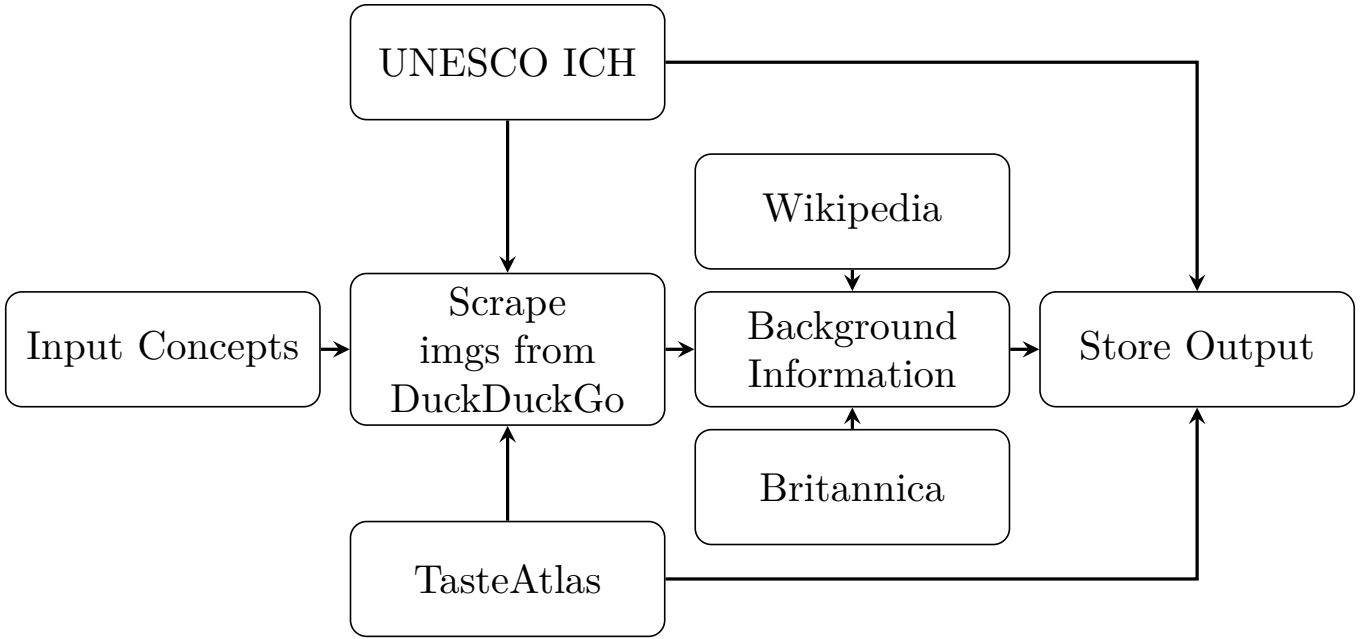
\begin{figure*}[t]
\centering
    \resizebox{\textwidth}{!}{%
    \begin{tikzpicture}[node distance=3cm, auto]
        \node[block] (input) {Input Concepts};
        \node[block, right of=input] (scrape) {Scrape imgs from DuckDuckGo};
        \node[block, right of=scrape] (background) {Background Information};
        \node[block, above of=background, node distance=1.2cm] (wikipedia) {Wikipedia};
        \node[block, below of=background, node distance=1.2cm] (britannica) {Britannica};
        \node[block, right of=background] (output) {Store Output};

        \node[block, below of=scrape, node distance=2cm] (tasteatlas) {TasteAtlas};
        \node[block, above of=scrape, node distance=2.5cm] (unesco) {UNESCO ICH};

        \draw[arrow] (input) -- (scrape);
        \draw[arrow] (scrape) -- (background);
        \draw[arrow] (wikipedia) -- (background);
        \draw[arrow] (britannica) -- (background);
        \draw[arrow] (background) -- (output);

        \draw[arrow] (tasteatlas) -- (scrape);
        \draw[arrow] (tasteatlas) -| (output);
        \draw[arrow] (unesco) -- (scrape);
        \draw[arrow] (unesco) -| (output);
    \end{tikzpicture}
    }
    \caption{Data collection pipeline with the base multimodal retrieval process.}
    \label{fig:pipeline}
\end{figure*}

\subsection{Ethical and Quality Considerations}
All data are sourced from publicly accessible and open-licensed platforms. Only images marked for reuse are included, and no personal or copyrighted materials are collected. A subset of data undergoes manual validation to ensure cultural accuracy and representational fairness. Future iterations will integrate automated filtering and expert curation mechanisms to enhance the dataset’s factual and cultural integrity.

\section{Dataset}
\subsection{Dataset Statistics}
By applying the automated pipeline to a list of Egyptian cultural concepts, we created \textit{EgMM-Corpus}, a multimodal dataset consisting of the following:
\begin{itemize}
    \item 313 cultural entities
    \item 3,000+ downloaded images
    \item Textual descriptions from Wikipedia and Britannica
    \item A Structured, folder-based output for each concept
\end{itemize}
Table~\ref{tab:corpus-statistics} presents the overall statistics of the \textit{EgMM-Corpus}.

\begin{table}[!ht]
\centering
\caption{EgMM-Corpus Statistics}
\label{tab:corpus-statistics}
\begin{tabular}{@{}lccc@{}}
\toprule
Dataset                      & Size (Samples) & Number of concepts \\ \midrule
\textbf{EgMM-Corpus}         & 3,130 & 313 \\ \bottomrule
\end{tabular}
\end{table}

\subsection{Dataset Composition}
The dataset is composed of three main categories with several predefined concepts under each category. The categories are as follows, and Table~\ref{tab:corpus-categories} shows the number of concepts for each category:
\begin{enumerate}
    \item Landmarks \& Monuments
    \item Food 
    \item Egyptian Folklor
\end{enumerate}

\begin{table}[!ht]
\centering
\caption{EgMM-Corpus Categories}
\label{tab:corpus-categories}
\begin{tabular}{@{}lccc@{}}
\toprule
Dataset                      & Landmarks \& Monuments & Food &  Folklor \\ \midrule
\textbf{EgMM-Corpus}         & 258 & 45 & 10 \\ \bottomrule
\end{tabular}
\end{table}
\section{Results}

We evaluate the zero-shot performance of CLIP (ViT-B/32) on the EgMM-Corpus to establish baseline performance on Egyptian cultural concepts. Our evaluation focuses on two key tasks: \textbf{Zero-shot classification} and \textbf{cross-modal retrieval}. Table~\ref{tab:results} presents the comprehensive results.

\subsection{Evaluation Setup}

The evaluation is conducted on a sample of Egyptian cultural concepts comprising 995 images sampled from our corpus. We employ a pure zero-shot setting where CLIP has no prior exposure to our dataset, testing its inherent understanding of Egyptian cultural content acquired during pre-training on web-scale data. For zero-shot classification, each image is matched against all concept names, and we report Top-1 and Top-5 accuracy. 
\begin{itemize}
    \item For retrieval, we adopt \textbf{Recall@K (R@K)}-a standard measure in cross-modal learning-evaluated in both directions:
    \begin{itemize}
        \item \textit{Image-to-Text (I2T)}: each image query is used to rank all textual descriptions. R@K represents the fraction of images for which the correct text appears among the top K retrieved results.
        \item \textit{Text-to-Image (T2I)}: each textual concept query is used to rank all images. R@K measures the percentage of times the correct image is found within the top K results.
    \end{itemize}
    \item We primarily report \textbf{R@1} (strict correctness) and \textbf{R@5} (top-5 relevance) to provide a comprehensive view of CLIP’s visual–textual alignment in a zero-shot setting.
\end{itemize}
\par Together, these metrics quantify both CLIP’s recognition accuracy and its retrieval alignment capability, revealing its ability to identify cultural elements effectively.

\subsection{Overall Performance}

\begin{table}[!ht]
\centering
\caption{CLIP performance on EgMM-Corpus (zero-shot classification and retrieval).}
\label{tab:results}
\begin{adjustbox}{width=\columnwidth}
\begin{tabular}{@{}lccccc@{}}
\toprule
\textbf{Model} & \textbf{Acc@1} & \textbf{Acc@5} & \textbf{I2T R@1} & \textbf{T2I R@1} & \textbf{I2T R@5} \\
\midrule
CLIP ViT-B/32 & 21.2\% & 36.4\% & 21.2\% & 18.7\% & 27.2\% \\
\bottomrule
\end{tabular}
\end{adjustbox}
\end{table}

CLIP achieves 21.2\% accuracy on zero-shot classification across all Egyptian cultural concepts, correctly identifying approximately one in five images. Top-5 accuracy reaches 36.4\%, indicating that the correct concept appears among the top five predictions in roughly one-third of cases. This gap between Top-1 and Top-5 performance suggests CLIP can establish partial semantic relationships but lacks precise cultural knowledge to confidently distinguish between similar Egyptian concepts.

Cross-modal retrieval results reveal consistent challenges in matching images to textual descriptions. Image-to-text retrieval  achieves 21.2\% R@1 and 27.2\% R@5, meaning that given an image of an Egyptian cultural site, CLIP retrieves the correct textual description only 21\% of the time as the top-ranked result. Text-to-image retrieval performs slightly worse at 18.7\% R@1, indicating even greater difficulty in matching concept names to their visual representations. These low retrieval scores have direct implications for practical applications such as cultural heritage search systems, digital museum platforms, and educational resources targeting Egyptian content.

\subsection{Per-Concept Analysis}

Analysis of per-concept performance reveals stark variations across the dataset. Five concepts achieve perfect accuracy (Luxor Temple, Cairo International Stadium, Edfu Temple, Coloured Canyon, Petrified Forest), suggesting these landmarks possess distinctive visual features that are well-represented in CLIP's training data. Conversely, ten concepts achieve zero accuracy, including significant cultural sites such as Bab~al-Nasr (Cairo), Khan~el-Khalili, and the Mortuary~Temple~of~Seti~I. This bimodal distribution-with numerous concepts at both performance extremes-indicates systematic gaps rather than uniform difficulty across Egyptian cultural content.

The results demonstrate that despite strong performance on standard vision--language benchmarks, CLIP exhibits significant limitations when evaluated on culturally specific content underrepresented in mainstream web data. These findings establish baseline performance metrics for future work and empirically justify the need for specialized datasets like EgMM-Corpus to enable the development of culturally aware multimodal models.

\section{Discussion}

EgMM-Corpus is a practical, purpose-built resource that targets a clear gap: multimodal, culturally grounded data for Egypt. Its primary strengths are a focused cultural scope (313 concepts, $\approx$3,000 images), multimodal alignment between images and curated textual grounding (Wikipedia/Britannica), and a simple, reproducible per-concept directory structure that makes the corpus immediately usable for retrieval, captioning, and evaluation tasks.

Practically, the dataset is well suited as a benchmark and a seed collection: it enables targeted tests of cultural robustness in existing vision--language models, supports few-shot and retrieval experiments, and provides human-curated examples that are valuable for qualitative analysis and prototype systems. The inclusion of food and folklore alongside landmarks broadens the cultural signal beyond purely architectural or tourist-centric images, which helps study social and intangible aspects of culture.

We are also careful about provenance and quality: sources were chosen for reliability, images were filtered for apparent reusability, and a manual validation pass was performed to reduce glaring errors. The documented pipeline and directory layout simplify extension and community contributions, so researchers can augment the corpus or adapt the pipeline for other MENA contexts.

At the same time, the current release has important limitations. Image counts per concept are modest, which limits large-scale supervised training and necessitates careful experimental design (e.g., few-shot or transfer settings).

Our evaluation reveals systematic performance gaps when CLIP is applied to Egyptian cultural content, with 21.2\% zero-shot accuracy demonstrating limited exposure to regional concepts despite impressive performance on Western-centric benchmarks The gap between Top-1 (21.2\%) and Top-5 (36.4\%) accuracy indicates CLIP recognizes visual similarity between Egyptian sites but lacks fine-grained discriminative knowledge to distinguish them confidently. Per-concept analysis reveals a bimodal distribution: perfect accuracy on internationally recognized landmarks (Luxor Temple, Edfu Temple, Coloured Canyon) versus zero accuracy on culturally significant but locally-known sites (Bab al-Nasr, Khan el-Khalili, Mortuary Temple of Seti I). This pattern reflects geographic and linguistic bias in CLIP's training data toward Western-produced content, with sites documented primarily in Arabic-language sources or specialized heritage databases remaining effectively invisible to the model.

\section{Conclusion and Future Work}
In this work, we presented \textit{EgMM-Corpus}, a multimodal vision–language dataset designed to represent Egyptian culture. Through a carefully designed pipeline, we could gather over 3,000 captioned images representing a wide range of domains within Egyptian culture.

\textit{EgMM-Corpus} helps bridge the gap in the culture datasets literature by providing a rich resource for evaluating and training vision-language models that can reason within an Egyptian cultural context.

We have also evaluated the zero-shot performance of CLIP on \textit{EgMM-Corpus}, which achieved 21.2\% Top-1 accuracy and 36.4\% Top-5 accuracy in classification, further indicating the significance of \textit{EgMM-Corpus} as a benchmark for developing culturally aware models.

Furthermore, the pipeline we designed establishes a scalable foundation for constructing multimodal datasets across diverse cultural domains. In future work, we aim to:
\begin{itemize}
    \item Expand coverage to additional countries and cultural categories (festivals, crafts, music),
    \item Incorporate large language models (LLMs) to generate culturally grounded visual question-answer pairs,
    \item Benchmark multimodal reasoning tasks using the resulting dataset.
\end{itemize}

\end{document}